\documentclass{article} % For LaTeX2e
\usepackage{mlgenx_conference,times}

\usepackage{amsmath,amsfonts,bm}

\def\eqref#1{equation~\ref{#1}}
\def\1{\bm{1}}

\DeclareMathAlphabet{\mathsfit}{\encodingdefault}{\sfdefault}{m}{sl}
\SetMathAlphabet{\mathsfit}{bold}{\encodingdefault}{\sfdefault}{bx}{n}

\usepackage[utf8]{inputenc}
\usepackage{hyperref}
\usepackage{url}
\usepackage{graphicx} % Required for inserting images
\usepackage{amsmath}
\usepackage{flafter} 
\usepackage{booktabs}
\usepackage{multirow}

\title{BioCOMPASS: Integrating Biomarkers into Transformer-Based Immunotherapy Response Prediction}

\author{Sayed Hashim, Frank Soboczenski \& Paul Cairns\\
University of York, UK\\
\texttt{\{sayed.hashim\}@york.ac.uk} \\
}

\mlgenxfinal % Uncomment for camera-ready version, but NOT for submission.
\begin{document}

\maketitle

\begin{abstract}
Datasets used in immunotherapy response prediction are typically small in size, as well as diverse in cancer type, drug administered, and sequencer used. Models often drop in performance when tested on patient cohorts that are not included in the training process. Recent work has shown that transformer-based models along with self-supervised learning show better generalisation performance than threshold-based biomarkers, but is still suboptimal. We present BioCOMPASS, an extension of a transformer-based model called COMPASS, that integrates biomarkers and treatment information to further improve its generalisability. Instead of feeding biomarker data as input, we built loss components to align them with the model's intermediate representations. We found that components such as treatment gating and pathway consistency loss improved generalisability when evaluated with Leave-one-cohort-out, Leave-one-cancer-type-out and Leave-one-treatment-out strategies. Results show that building components that exploit biomarker and treatment information can help in generalisability of immunotherapy response prediction. Careful curation of additional components that leverage complementary clinical information and domain knowledge represents a promising direction for future research.
\end{abstract}

\section{Introduction}
The immune system is responsible for the management of cancer and the identification of neoantigens produced by tumour cells that can trigger cellular immune responses \citep{grivennikov_immunity_2010}. However, tumour cells have devised ways to avoid immune surveillance \citep{rabinovich_immunosuppressive_2007}. To tackle this challenge, cancer immunotherapy emerged with the objective of reinstating the immune system's capacity to identify and destroy cancer cells \citep{li_informing_2024}. Although immunotherapy has improved the prognosis for patients, its success is limited to a select, unpredictable fraction of individuals diagnosed with cancer \citep{drake_breathing_2014}. Thus, accurate characterisation of the tumour microenvironment (TME) in a patient with the ability to anticipate responses to immunotherapy is critical to enhance the efficiency of immunotherapy treatment strategies \citep{li_informing_2024}.

Existing methods for predicting the efficacy of immunotherapy are predominantly dependent on specific biomarkers, including the level of immune cell infiltration \citep{simoni_bystander_2018}, the expression levels of programmed death 1 (PD-1) and programmed death-ligand 1 (PD-L1) \citep{garon_pembrolizumab_2015}, the expression of the cytotoxic T lymphocyte-associated protein 4 (CTLA-4) \citep{leach_enhancement_1996}, as well as tumour mutational burden (TMB) \citep{rizvi_mutational_2015}. However, current clinical methodologies that rely on threshold-based approaches are often inadequate \citep{li_informing_2024}. Many machine learning (ML)-based methods have been proposed to estimate biomarkers and treatment outcomes \citep{li_informing_2024}. These models face challenges when tasked with new data that they were not previously trained on. When evaluated on new datasets, their performance tends to be mediocre or even inadequate, highlighting a gap in their ability to generalise \citep{li_informing_2024}. 
% In addition, marker genes identified based on ML methods tend to differ across various studies and may not be impactful when applied to external datasets \citep{li_informing_2024}. 

A recent work called COMPASS \citep{shen_generalizable_2025} used self-supervised learning (SSL) with a transformer-based encoder and a biologically grounded concept bottleneck layer to improve performance across cancer types and treatments. COMPASS is pre-trained on gene expression data from 33 types of cancers using a triplet loss based SSL method. Pre-training improves its generalisability, while the concept bottleneck enables interpretability. COMPASS is fine-tuned on clinical cohorts to predict immunotherapy response. In COMPASS, patient embeddings produced from the encoder are passed onto a concept bottleneck layer to generate scores for 44 biological concepts such as genome integrity, cell proliferation and immune checkpoint for each tumour. These are then passed onto a classifier module to generate treatment response probabilities. 

The authors of COMPASS used Leave-one-cohort-out (LOCO) strategy to evaluate its generalisability. In this setting, all cohorts except one are used for training, and the left-out cohort is used for testing. Although the generalisability of COMPASS in this setting is better than methods that use single biomarkers, such as the expression of PD-1 or PDL-1, it is still suboptimal. For instance, the accuracy of COMPASS is about 65\% across small cohorts (sample size less than 20) in LOCO setting as reported in their publication. Moreover, COMPASS does not make use of treatment information or external biomarkers. As treatment response could vary based on the treatment type, it is vital to feed this information into the model. COMPASS also does not have a way to validate its concepts with external biomarkers during training.

We present BioCOMPASS, a modified version of COMPASS that integrates external biomarkers and treatment information using components including treatment gating, concept alignment, pathway consistency, and auxiliary multi-task learning into COMPASS. Our contributions are the following.
\begin{enumerate}
    \item A treatment gating layer to feed information about the target of the treatment (eg. PD-1, CTLA-4, combination) into the model so that it produces treatment-aware concepts.
    \item Alignment between known external biomarker scores and concept scores produced by the model to ensure that concept scores are validated against biomarker scores during training.
    \item Pathway consistency loss to ensure that embeddings produced by the model contain pathway relevant information and are biologically grounded.
\end{enumerate}

In short, BioCOMPASS is an extension of COMPASS that exploits treatment information as well as external biomarker and pathway scores in order to make the model treatment-aware, biologically grounded, and thus more generalisable. This work also shows that rather than feeding clinical and biomarker data as input to the model, aligning the intermediate latent representations using them could be a good avenue to pursue in general medical applications, especially if such data is only available during training and not inference.

\section{Methods}
\subsection{Data}
\label{subsec:data}
The authors of COMPASS finetuned the model on a total of 16 cohorts. However, COMPASS does not provide access to these datasets; rather, they list the original publications of the cohorts, and access to these datasets needs to be requested from the original publications. Due to difficulties in accessing them, we visited the CRI iAtlas portal \citep{eddy_cri_2020}, which contains preprocessed gene expression data, biomarkers, and treatment information for 8 out of the 16 immunotherapy cohorts. We downloaded data for 8 of these cohorts from CRI iAtlas. Due to issues in accessing data of all cohorts in COMPASS, we reproduced results for COMPASS using the 8 cohorts we could obtain and used the pretrained model from COMPASS with its default hyperparameters to finetune.

Information on the sample size of the cohorts, the drug used in them, and their publication is given in the Appendix \ref{app:cohorts}. The gene expression data was already normalised to Transcripts Per Million (TPM) units. A binary responder label derived from the labels based on response evaluation criteria in solid tumours (RECIST) was used for classification. BioCOMPASS is finetuned to predict this binary label from gene expression data.

\subsection{Model Architecture}
We added biomarker and clinical components on top of the COMPASS architecture to build BioCOMPASS, as shown in Figure \ref{fig:arch}. The formulae for the components are in Appendix \ref{app:model}. The implementation is available at \href{https://github.com/hashimsayed0/BioCOMPASS}{https://github.com/hashimsayed0/BioCOMPASS}.

\begin{figure}[]
    \centering
    % \includesvg[width=1.0\linewidth]{images/biocompass_architecture.svg}
    \includegraphics[width=\textwidth]{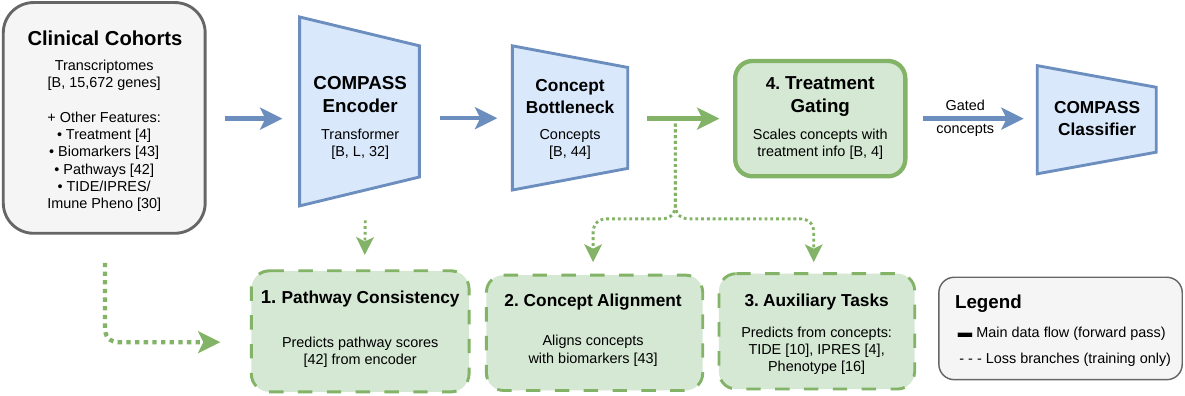}
    \caption{BioCOMPASS architecture: Gene expression data is first fed into the COMPASS encoder to generate embeddings. Minimising the pathway consistency loss makes sure that the pathway scores predicted from embeddings are aligned with external pathway scores. The embeddings are then fed into the COMPASS concept bottleneck to generate 44 biological concepts. These are aligned with cell-type biomarker scores using the concept alignment objective. They are also used to predict immunotherapy response prediction biomarkers such as TIDE \& IPRES and other immune phenotypes. The concepts are also scaled based on the specific treatment type using the treatment gating module. The scaled concepts are then used to predict response using a classifier head. Components from COMPASS are in blue colour while BioCOMPASS components are in green.}
    \label{fig:arch}
\end{figure}

\textbf{Pathway Consistency}: An auxiliary head containing fully connected layers is trained to predict external pathway activity scores (42 CTLA-4/PD-1 pathway features) from gene embeddings by minimising the mean-squared error (MSE) loss between them. This encourages the encoder to learn representations that are pathway relevant and not cohort-specific noise.

\textbf{Concept Alignment}: Concept alignment involves making the model align learnt concepts (eg. plasma cell, cytotoxic T-cell) and external biomarker scores (e.g., cell type abundances). A projection layer is used to bring them to the same latent dimension. The distance between concept projections and biomarker scores is then minimised.

\textbf{Auxiliary Tasks}: This component involves multi-task learning by predicting established biomarker scores (TIDE, IPRES, and immune phenotypes) from concepts alongside response prediction. This is done through separate decoder heads attached to the concept bottleneck layer. MSE loss between predictions from the auxiliary decoder heads and the actual scores is minimised. 

\textbf{Treatment Gating}: Treatment gating scales biological concepts based on the target of immunotherapy treatment (PD-1, CTLA-4, combination). Treatment indicators are embedded and passed through a gating network to compute gate weights, which are then multiplied with the concepts. This allows the model to adaptively focus on concepts relevant to each treatment type and thus integrate treatment information into the concepts. 

\section{Results}
\begin{table}[!htbp]
\centering
\caption{LOCO validation of COMPASS (C) and BioCOMPASS (BC). This table shows the average performance (in \%) across all left-out cohorts. The first two rows show results averaged across all 8 test cohorts; the ones below show the same across 4 small cohorts (less than 50 samples) and 4 large cohorts (more than 50 samples). The 95\% confidence intervals (CI) show variation across 4 seeds.\newline}
\label{tab:loco}

\begin{tabular}{@{}lllllll@{}}
\toprule
\textbf{Cohorts}                                                                    & \textbf{Model} & \textbf{Accuracy}     & \textbf{ROC-AUC}     & \textbf{F1}      & \textbf{Precision}    & \textbf{Recall}   \\ \midrule
\multirow{2}{*}{All}                                                                & C              & 63.10 ± 5.43          & 70.99 ± 2.88          & 46.65 ± 2.15          & 48.33 ± 6.24          & 56.93 ± 6.51           \\
                                                                                    & BC             & \textbf{70.00 ± 1.76} & \textbf{73.58 ± 1.29} & \textbf{54.01 ± 2.81} & \textbf{56.00 ± 2.64} & \textbf{58.55 ± 6.55}  \\
\multirow{2}{*}{\begin{tabular}[c]{@{}l@{}}Small\\ (\textless{}50)\end{tabular}}    & C              & 63.03 ± 6.15          & 72.05 ± 4.30          & 39.55 ± 3.98          & 40.82 ± 8.86          & 51.88 ± 3.91           \\
                                                                                    & BC             & \textbf{69.77 ± 1.86} & \textbf{74.44 ± 2.46} & \textbf{52.93 ± 1.67} & \textbf{52.51 ± 4.30} & \textbf{61.95 ± 6.96}  \\
\multirow{2}{*}{\begin{tabular}[c]{@{}l@{}}Large\\ (\textgreater{}50)\end{tabular}} & C              & 63.18 ± 6.74          & 69.94 ± 2.75          & 53.74 ± 6.33          & 55.84 ± 6.25          & \textbf{61.98 ± 15.36} \\
                                                                                    & BC             & \textbf{70.24 ± 4.84} & \textbf{72.72 ± 0.80} & \textbf{55.08 ± 3.96} & \textbf{59.48 ± 5.83} & 55.14 ± 9.82           \\ \bottomrule
\end{tabular}
\end{table}

We ran experiments to compare BioCOMPASS with COMPASS. We initialised both models with weights from the COMPASS model pretrained on The Cancer Genome Atlas (TCGA) data \citep{weinstein_cancer_2013} and finetuned them in PFT (partial fine tuning) mode. This mode involves freezing the encoder and only training the concept bottleneck and classifier head. In BioCOMPASS, biomarker data is used during training, but is not required during inference. Each run was done 4 times with 4 different seeds to ensure robustness of results. 
% The difference in seeds could result in different convergence paths during training. 

\begin{figure}[htbp]
    \centering
    \includegraphics[width=1.0\linewidth]{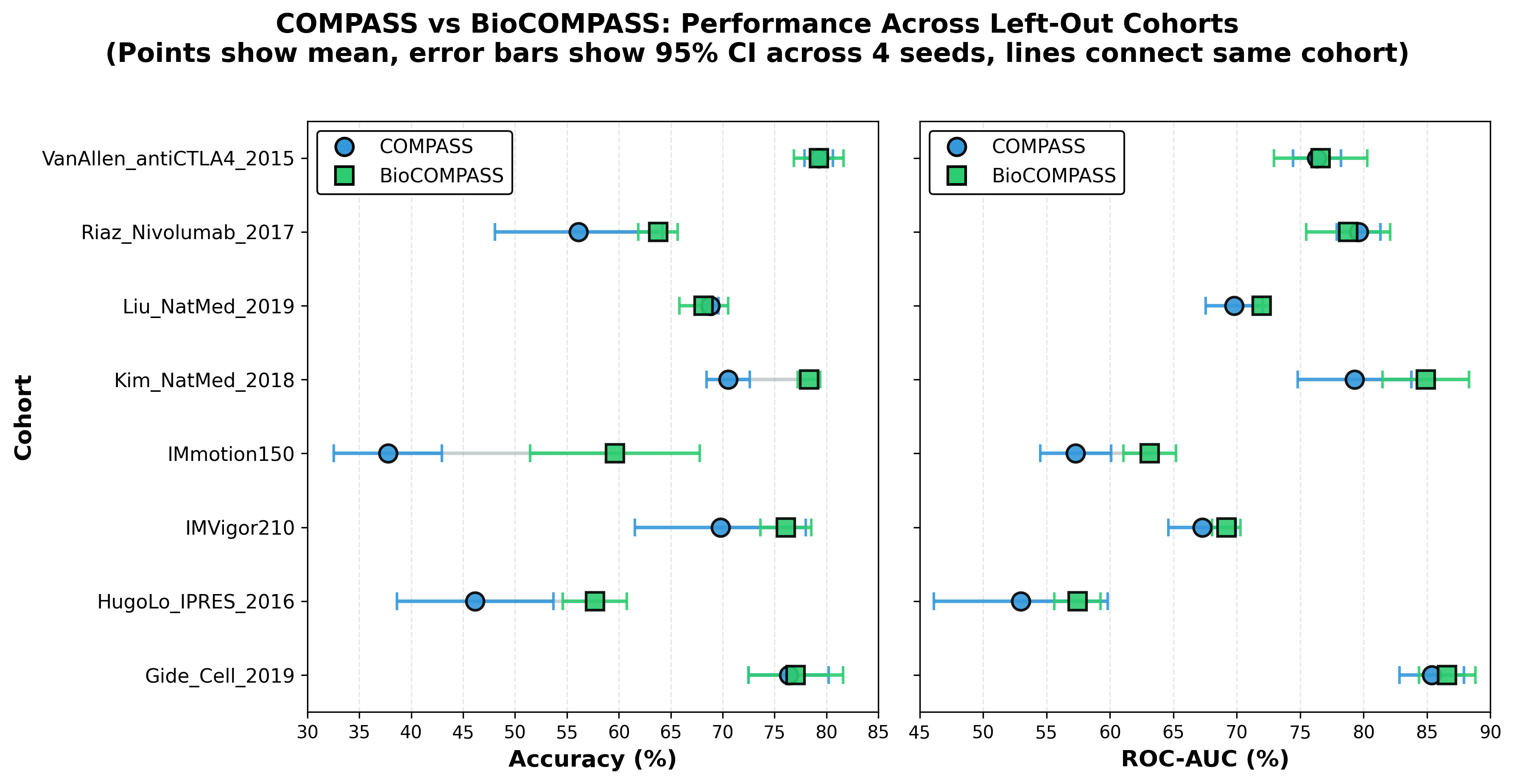}
    \caption{Points show mean performance across four random seeds with 95\% CI error bars. Circles: COMPASS; Squares: BioCOMPASS. BioCOMPASS shows consistent improvements in accuracy and ROC-AUC across most cohorts. Please note that the subplots have different scales on x-axis.}
    \label{fig:cohort}
\end{figure}

Table \ref{tab:loco} shows the average performance on the left-out cohort across all 8 cohorts in LOCO setting across 4 seeds. BioCOMPASS excels over COMPASS in all metrics and settings except recall of large cohorts, which could be because BioCOMPASS might be more conservative in its predictions, evident from its higher precision. However, higher ROC-AUC and F1 score show its superior performance. Figure \ref{fig:cohort} shows the performance on each left-out cohort across 4 seeds. The metrics for COMPASS are obtained by reproducing on the 8 cohorts we could obtain and not all 16 cohorts as described in Section \ref{subsec:data}. Ablation studies showed that treatment gating is the most influential component, followed by pathway consistency. Results of ablation study are given in Appendix \ref{app:ablation}. To further evaluate its generalisability, we also ran experiments in Leave-one-cancer-type-out (LOCTO) and Leave-one-treatment-out (LOTO) settings. BioCOMPASS excels over COMPASS in those settings as well as shown in Appendix \ref{app:addgen}. We also trained logistic regression on biomarker-based baseline methods as well gene expression data. But they do not generalise well as shown in Appendix \ref{app:baseline}. 

\section{Discussion}
Results show that aligning representations using biomarkers and treatment information in the training process of BioCOMPASS helps it show better generalisation performance over COMPASS in LOCO, LOCTO and LOTO settings. Although the training process of BioCOMPASS requires biomarkers, only treatment information is required during inference. A possible future direction is to exploit domain information about concepts and treatment, such as by using biomedical text mining or biological knowledge graphs, to improve the latent representations. In conclusion, integrating richer clinical and domain knowledge into transformer-based architectures, through informed attention mechanisms or structured latent representation, could further enhance representation learning, robustness, and generalisation across heterogeneous patient cohorts.

% TODO: Uncomment for camera ready.
\section{Acknowledgments}
This work was supported by UKRI AI Centre for Doctoral Training in Safe Artificial Intelligence Systems (SAINTS) (EP/Y030540/1).

\bibliography{references}

@article{mariathasan_tgfb_2018,
	title = {{TGFB} attenuates tumour response to {PD}-{L1} blockade by contributing to exclusion of {T} cells},
	volume = {554},
	copyright = {2018 Macmillan Publishers Limited, part of Springer Nature. All rights reserved.},
	issn = {1476-4687},
	url = {https://www.nature.com/articles/nature25501},
	doi = {10.1038/nature25501},
	language = {en},
	number = {7693},
	urldate = {2026-02-03},
	journal = {Nature},
	author = {Mariathasan, Sanjeev and Turley, Shannon J. and Nickles, Dorothee and Castiglioni, Alessandra and Yuen, Kobe and Wang, Yulei and Kadel III, Edward E. and Koeppen, Hartmut and Astarita, Jillian L. and Cubas, Rafael and Jhunjhunwala, Suchit and Banchereau, Romain and Yang, Yagai and Guan, Yinghui and Chalouni, Cecile and Ziai, James and Şenbabaoğlu, Yasin and Santoro, Stephen and Sheinson, Daniel and Hung, Jeffrey and Giltnane, Jennifer M. and Pierce, Andrew A. and Mesh, Kathryn and Lianoglou, Steve and Riegler, Johannes and Carano, Richard A. D. and Eriksson, Pontus and Höglund, Mattias and Somarriba, Loan and Halligan, Daniel L. and van der Heijden, Michiel S. and Loriot, Yohann and Rosenberg, Jonathan E. and Fong, Lawrence and Mellman, Ira and Chen, Daniel S. and Green, Marjorie and Derleth, Christina and Fine, Gregg D. and Hegde, Priti S. and Bourgon, Richard and Powles, Thomas},
	month = feb,
	year = {2018},
	pages = {544--548},
}

@article{ayers_ifn-grelated_2017,
	title = {{IFN}-\textbf{{G}}–related {mRNA} profile predicts clinical response to {PD}-1 blockade},
	volume = {127},
	issn = {0021-9738},
	url = {https://www.jci.org/articles/view/91190},
	doi = {10.1172/JCI91190},
	language = {en},
	number = {8},
	urldate = {2026-03-16},
	journal = {The Journal of Clinical Investigation},
	author = {Ayers, Mark and Lunceford, Jared and Nebozhyn, Michael and Murphy, Erin and Loboda, Andrey and Kaufman, David R. and Albright, Andrew and Cheng, Jonathan D. and Kang, S. Peter and Shankaran, Veena and Piha-Paul, Sarina A. and Yearley, Jennifer and Seiwert, Tanguy Y. and Ribas, Antoni and McClanahan, Terrill K.},
	month = aug,
	year = {2017},
	pmid = {0},
	pages = {2930--2940},
}

@article{huang_single_2019,
	title = {A single dose of neoadjuvant {PD}-1 blockade predicts clinical outcomes in resectable melanoma},
	volume = {25},
	copyright = {2019 The Author(s), under exclusive licence to Springer Nature America, Inc.},
	issn = {1546-170X},
	url = {https://www.nature.com/articles/s41591-019-0357-y},
	doi = {10.1038/s41591-019-0357-y},
	language = {en},
	number = {3},
	urldate = {2026-03-16},
	journal = {Nature Medicine},
	author = {Huang, Alexander C. and Orlowski, Robert J. and Xu, Xiaowei and Mick, Rosemarie and George, Sangeeth M. and Yan, Patrick K. and Manne, Sasikanth and Kraya, Adam A. and Wubbenhorst, Bradley and Dorfman, Liza and D’Andrea, Kurt and Wenz, Brandon M. and Liu, Shujing and Chilukuri, Lakshmi and Kozlov, Andrew and Carberry, Mary and Giles, Lydia and Kier, Melanie W. and Quagliarello, Felix and McGettigan, Suzanne and Kreider, Kristin and Annamalai, Lakshmanan and Zhao, Qing and Mogg, Robin and Xu, Wei and Blumenschein, Wendy M. and Yearley, Jennifer H. and Linette, Gerald P. and Amaravadi, Ravi K. and Schuchter, Lynn M. and Herati, Ramin S. and Bengsch, Bertram and Nathanson, Katherine L. and Farwell, Michael D. and Karakousis, Giorgos C. and Wherry, E. John and Mitchell, Tara C.},
	month = mar,
	year = {2019},
	pages = {454--461},
}

@article{freeman_combined_2022,
	title = {Combined tumor and immune signals from genomes or transcriptomes predict outcomes of checkpoint inhibition in melanoma},
	volume = {3},
	issn = {2666-3791},
	url = {https://www.sciencedirect.com/science/article/pii/S2666379121003773},
	doi = {10.1016/j.xcrm.2021.100500},
	number = {2},
	urldate = {2026-03-16},
	journal = {Cell Reports Medicine},
	author = {Freeman, Samuel S. and Sade-Feldman, Moshe and Kim, Jaegil and Stewart, Chip and Gonye, Anna L. K. and Ravi, Arvind and Arniella, Monica B. and Gushterova, Irena and LaSalle, Thomas J. and Blaum, Emily M. and Yizhak, Keren and Frederick, Dennie T. and Sharova, Tatyana and Leshchiner, Ignaty and Elagina, Liudmila and Spiro, Oliver G. and Livitz, Dimitri and Rosebrock, Daniel and Aguet, François and Carrot-Zhang, Jian and Ha, Gavin and Lin, Ziao and Chen, Jonathan H. and Barzily-Rokni, Michal and Hammond, Marc R. and Vitzthum von Eckstaedt, Hans C. and Blackmon, Shauna M. and Jiao, Yunxin J. and Gabriel, Stacey and Lawrence, Donald P. and Duncan, Lyn M. and Stemmer-Rachamimov, Anat O. and Wargo, Jennifer A. and Flaherty, Keith T. and Sullivan, Ryan J. and Boland, Genevieve M. and Meyerson, Matthew and Getz, Gad and Hacohen, Nir},
	month = feb,
	year = {2022},
	pages = {100500},
}

@article{auslander_robust_2018,
	title = {Robust prediction of response to immune checkpoint blockade therapy in metastatic melanoma},
	volume = {24},
	copyright = {2018 The Author(s)},
	issn = {1546-170X},
	url = {https://www.nature.com/articles/s41591-018-0157-9},
	doi = {10.1038/s41591-018-0157-9},
	language = {en},
	number = {10},
	urldate = {2026-03-16},
	journal = {Nature Medicine},
	author = {Auslander, Noam and Zhang, Gao and Lee, Joo Sang and Frederick, Dennie T. and Miao, Benchun and Moll, Tabea and Tian, Tian and Wei, Zhi and Madan, Sanna and Sullivan, Ryan J. and Boland, Genevieve and Flaherty, Keith and Herlyn, Meenhard and Ruppin, Eytan},
	month = oct,
	year = {2018},
	pages = {1545--1549},
}

@article{giordano_molecular_2015,
	title = {Molecular profiling of {CD8} {T} cells in autochthonous melanoma identifies {Maf} as driver of exhaustion},
	volume = {34},
	issn = {1460-2075},
	url = {https://doi.org/10.15252/embj.201490786},
	doi = {10.15252/embj.201490786},
	language = {en},
	number = {15},
	urldate = {2026-03-16},
	journal = {The EMBO Journal},
	author = {Giordano, Marilyn and Henin, Coralie and Maurizio, Julien and Imbratta, Claire and Bourdely, Pierre and Buferne, Michel and Baitsch, Lukas and Vanhille, Laurent and Sieweke, Michael H. and Speiser, Daniel E. and Auphan‐Anezin, Nathalie and Schmitt‐Verhulst, Anne‐Marie and Verdeil, Grégory},
	month = aug,
	year = {2015},
	pages = {2042--2058},
}

@article{nurmik_search_2020,
	title = {In search of definitions: {Cancer}‐associated fibroblasts and their markers},
	volume = {146},
	issn = {0020-7136, 1097-0215},
	shorttitle = {In search of definitions},
	url = {https://onlinelibrary.wiley.com/doi/10.1002/ijc.32193},
	doi = {10.1002/ijc.32193},
	language = {en},
	number = {4},
	urldate = {2026-03-16},
	journal = {International Journal of Cancer},
	author = {Nurmik, Martin and Ullmann, Pit and Rodriguez, Fabien and Haan, Serge and Letellier, Elisabeth},
	month = feb,
	year = {2020},
	pages = {895--905},
}

@article{fehrenbacher_atezolizumab_2016,
	title = {Atezolizumab versus docetaxel for patients with previously treated non-small-cell lung cancer ({POPLAR}): a multicentre, open-label, phase 2 randomised controlled trial},
	volume = {387},
	issn = {0140-6736},
	shorttitle = {Atezolizumab versus docetaxel for patients with previously treated non-small-cell lung cancer ({POPLAR})},
	url = {https://www.sciencedirect.com/science/article/pii/S0140673616005870},
	doi = {10.1016/S0140-6736(16)00587-0},
	number = {10030},
	urldate = {2026-03-16},
	journal = {The Lancet},
	author = {Fehrenbacher, Louis and Spira, Alexander and Ballinger, Marcus and Kowanetz, Marcin and Vansteenkiste, Johan and Mazieres, Julien and Park, Keunchil and Smith, David and Artal-Cortes, Angel and Lewanski, Conrad and Braiteh, Fadi and Waterkamp, Daniel and He, Pei and Zou, Wei and Chen, Daniel S and Yi, Jing and Sandler, Alan and Rittmeyer, Achim},
	month = apr,
	year = {2016},
	pages = {1837--1846},
}

@article{roh_integrated_2017,
	title = {Integrated molecular analysis of tumor biopsies on sequential {CTLA}-4 and {PD}-1 blockade reveals markers of response and resistance},
	volume = {9},
	copyright = {http://www.sciencemag.org/about/science-licenses-journal-article-reuse},
	issn = {1946-6234, 1946-6242},
	url = {https://www.science.org/doi/10.1126/scitranslmed.aah3560},
	doi = {10.1126/scitranslmed.aah3560},
	language = {en},
	number = {379},
	urldate = {2026-03-16},
	journal = {Science Translational Medicine},
	author = {Roh, Whijae and Chen, Pei-Ling and Reuben, Alexandre and Spencer, Christine N. and Prieto, Peter A. and Miller, John P. and Gopalakrishnan, Vancheswaran and Wang, Feng and Cooper, Zachary A. and Reddy, Sangeetha M. and Gumbs, Curtis and Little, Latasha and Chang, Qing and Chen, Wei-Shen and Wani, Khalida and De Macedo, Mariana Petaccia and Chen, Eveline and Austin-Breneman, Jacob L. and Jiang, Hong and Roszik, Jason and Tetzlaff, Michael T. and Davies, Michael A. and Gershenwald, Jeffrey E. and Tawbi, Hussein and Lazar, Alexander J. and Hwu, Patrick and Hwu, Wen-Jen and Diab, Adi and Glitza, Isabella C. and Patel, Sapna P. and Woodman, Scott E. and Amaria, Rodabe N. and Prieto, Victor G. and Hu, Jianhua and Sharma, Padmanee and Allison, James P. and Chin, Lynda and Zhang, Jianhua and Wargo, Jennifer A. and Futreal, P. Andrew},
	month = mar,
	year = {2017},
	pages = {eaah3560},
}

@article{rooney_molecular_2015,
	title = {Molecular and {Genetic} {Properties} of {Tumors} {Associated} with {Local} {Immune} {Cytolytic} {Activity}},
	volume = {160},
	issn = {0092-8674},
	url = {https://www.sciencedirect.com/science/article/pii/S0092867414016390},
	doi = {10.1016/j.cell.2014.12.033},
	number = {1},
	urldate = {2026-03-16},
	journal = {Cell},
	author = {Rooney, Michael S. and Shukla, Sachet A. and Wu, Catherine J. and Getz, Gad and Hacohen, Nir},
	month = jan,
	year = {2015},
	pages = {48--61},
}

@article{davoli_tumor_2017,
	title = {Tumor aneuploidy correlates with markers of immune evasion and with reduced response to immunotherapy},
	volume = {355},
	copyright = {http://www.sciencemag.org/about/science-licenses-journal-article-reuse},
	issn = {0036-8075, 1095-9203},
	url = {https://www.science.org/doi/10.1126/science.aaf8399},
	doi = {10.1126/science.aaf8399},
	language = {en},
	number = {6322},
	urldate = {2026-03-16},
	journal = {Science},
	author = {Davoli, Teresa and Uno, Hajime and Wooten, Eric C. and Elledge, Stephen J.},
	month = jan,
	year = {2017},
	pages = {eaaf8399},
}

@article{joyce_t_2015,
	title = {T cell exclusion, immune privilege, and the tumor microenvironment},
	volume = {348},
	copyright = {http://www.sciencemag.org/about/science-licenses-journal-article-reuse},
	issn = {0036-8075, 1095-9203},
	url = {https://www.science.org/doi/10.1126/science.aaa6204},
	doi = {10.1126/science.aaa6204},
	language = {en},
	number = {6230},
	urldate = {2026-03-16},
	journal = {Science},
	author = {Joyce, Johanna A. and Fearon, Douglas T.},
	month = apr,
	year = {2015},
	pages = {74--80},
}

@article{jiang_signatures_2018,
	title = {Signatures of {T} cell dysfunction and exclusion predict cancer immunotherapy response},
	volume = {24},
	copyright = {2018 The Author(s)},
	issn = {1546-170X},
	url = {https://www.nature.com/articles/s41591-018-0136-1},
	doi = {10.1038/s41591-018-0136-1},
	language = {en},
	number = {10},
	urldate = {2026-03-16},
	journal = {Nature Medicine},
	author = {Jiang, Peng and Gu, Shengqing and Pan, Deng and Fu, Jingxin and Sahu, Avinash and Hu, Xihao and Li, Ziyi and Traugh, Nicole and Bu, Xia and Li, Bo and Liu, Jun and Freeman, Gordon J. and Brown, Myles A. and Wucherpfennig, Kai W. and Liu, X. Shirley},
	month = oct,
	year = {2018},
	pages = {1550--1558},
}

@article{chen_analysis_2016,
	title = {Analysis of {Immune} {Signatures} in {Longitudinal} {Tumor} {Samples} {Yields} {Insight} into {Biomarkers} of {Response} and {Mechanisms} of {Resistance} to {Immune} {Checkpoint} {Blockade}},
	volume = {6},
	issn = {2159-8274, 2159-8290},
	url = {https://aacrjournals.org/cancerdiscovery/article/6/8/827/5660/Analysis-of-Immune-Signatures-in-Longitudinal},
	doi = {10.1158/2159-8290.CD-15-1545},
	language = {en},
	number = {8},
	urldate = {2026-03-16},
	journal = {Cancer Discovery},
	author = {Chen, Pei-Ling and Roh, Whijae and Reuben, Alexandre and Cooper, Zachary A. and Spencer, Christine N. and Prieto, Peter A. and Miller, John P. and Bassett, Roland L. and Gopalakrishnan, Vancheswaran and Wani, Khalida and De Macedo, Mariana Petaccia and Austin-Breneman, Jacob L. and Jiang, Hong and Chang, Qing and Reddy, Sangeetha M. and Chen, Wei-Shen and Tetzlaff, Michael T. and Broaddus, Russell J. and Davies, Michael A. and Gershenwald, Jeffrey E. and Haydu, Lauren and Lazar, Alexander J. and Patel, Sapna P. and Hwu, Patrick and Hwu, Wen-Jen and Diab, Adi and Glitza, Isabella C. and Woodman, Scott E. and Vence, Luis M. and Wistuba, Ignacio I. and Amaria, Rodabe N. and Kwong, Lawrence N. and Prieto, Victor and Davis, R. Eric and Ma, Wencai and Overwijk, Willem W. and Sharpe, Arlene H. and Hu, Jianhua and Futreal, P. Andrew and Blando, Jorge and Sharma, Padmanee and Allison, James P. and Chin, Lynda and Wargo, Jennifer A.},
	month = aug,
	year = {2016},
	pages = {827--837},
}

@article{wu_prediction_2022,
	title = {Prediction of biomarkers and therapeutic combinations for anti-{PD}-1 immunotherapy using the global gene network association},
	volume = {13},
	copyright = {2022 The Author(s)},
	issn = {2041-1723},
	url = {https://www.nature.com/articles/s41467-021-27651-4},
	doi = {10.1038/s41467-021-27651-4},
	language = {en},
	number = {1},
	urldate = {2026-03-16},
	journal = {Nature Communications},
	author = {Wu, Chia-Chin and Wang, Y. Alan and Livingston, J. Andrew and Zhang, Jianhua and Futreal, P. Andrew},
	month = jan,
	year = {2022},
	pages = {42},
}

@article{cristescu_pan-tumor_2018,
	title = {Pan-tumor genomic biomarkers for {PD}-1 checkpoint blockade–based immunotherapy},
	volume = {362},
	issn = {0036-8075, 1095-9203},
	url = {https://www.science.org/doi/10.1126/science.aar3593},
	doi = {10.1126/science.aar3593},
	language = {en},
	number = {6411},
	urldate = {2026-03-16},
	journal = {Science},
	author = {Cristescu, Razvan and Mogg, Robin and Ayers, Mark and Albright, Andrew and Murphy, Erin and Yearley, Jennifer and Sher, Xinwei and Liu, Xiao Qiao and Lu, Hongchao and Nebozhyn, Michael and Zhang, Chunsheng and Lunceford, Jared K. and Joe, Andrew and Cheng, Jonathan and Webber, Andrea L. and Ibrahim, Nageatte and Plimack, Elizabeth R. and Ott, Patrick A. and Seiwert, Tanguy Y. and Ribas, Antoni and McClanahan, Terrill K. and Tomassini, Joanne E. and Loboda, Andrey and Kaufman, David},
	month = oct,
	year = {2018},
	pages = {eaar3593},
}

@article{messina_12-chemokine_2012,
	title = {12-{Chemokine} {Gene} {Signature} {Identifies} {Lymph} {Node}-like {Structures} in {Melanoma}: {Potential} for {Patient} {Selection} for {Immunotherapy}?},
	volume = {2},
	copyright = {2012 The Author(s)},
	issn = {2045-2322},
	shorttitle = {12-{Chemokine} {Gene} {Signature} {Identifies} {Lymph} {Node}-like {Structures} in {Melanoma}},
	url = {https://www.nature.com/articles/srep00765},
	doi = {10.1038/srep00765},
	language = {en},
	number = {1},
	urldate = {2026-03-16},
	journal = {Scientific Reports},
	author = {Messina, Jane L. and Fenstermacher, David A. and Eschrich, Steven and Qu, Xiaotao and Berglund, Anders E. and Lloyd, Mark C. and Schell, Michael J. and Sondak, Vernon K. and Weber, Jeffrey S. and Mulé, James J.},
	month = oct,
	year = {2012},
	pages = {765},
}

@article{hugo_genomic_2016,
	title = {Genomic and {Transcriptomic} {Features} of {Response} to {Anti}-{PD}-1 {Therapy} in {Metastatic} {Melanoma}},
	volume = {165},
	issn = {0092-8674},
	url = {https://www.sciencedirect.com/science/article/pii/S009286741630215X},
	doi = {10.1016/j.cell.2016.02.065},
	number = {1},
	urldate = {2026-02-03},
	journal = {Cell},
	author = {Hugo, Willy and Zaretsky, Jesse M. and Sun, Lu and Song, Chunying and Moreno, Blanca Homet and Hu-Lieskovan, Siwen and Berent-Maoz, Beata and Pang, Jia and Chmielowski, Bartosz and Cherry, Grace and Seja, Elizabeth and Lomeli, Shirley and Kong, Xiangju and Kelley, Mark C. and Sosman, Jeffrey A. and Johnson, Douglas B. and Ribas, Antoni and Lo, Roger S.},
	month = mar,
	year = {2016},
	pages = {35--44},
}

@article{gide_distinct_2019,
	title = {Distinct {Immune} {Cell} {Populations} {Define} {Response} to {Anti}-{PD}-1 {Monotherapy} and {Anti}-{PD}-1/{Anti}-{CTLA}-4 {Combined} {Therapy}},
	volume = {35},
	issn = {15356108},
	url = {https://linkinghub.elsevier.com/retrieve/pii/S1535610819300376},
	doi = {10.1016/j.ccell.2019.01.003},
	language = {en},
	number = {2},
	urldate = {2026-02-03},
	journal = {Cancer Cell},
	author = {Gide, Tuba N. and Quek, Camelia and Menzies, Alexander M. and Tasker, Annie T. and Shang, Ping and Holst, Jeff and Madore, Jason and Lim, Su Yin and Velickovic, Rebecca and Wongchenko, Matthew and Yan, Yibing and Lo, Serigne and Carlino, Matteo S. and Guminski, Alexander and Saw, Robyn P.M. and Pang, Angel and McGuire, Helen M. and Palendira, Umaimainthan and Thompson, John F. and Rizos, Helen and Silva, Ines Pires Da and Batten, Marcel and Scolyer, Richard A. and Long, Georgina V. and Wilmott, James S.},
	month = feb,
	year = {2019},
	pages = {238--255.e6},
}

@article{kim_comprehensive_2018,
	title = {Comprehensive molecular characterization of clinical responses to {PD}-1 inhibition in metastatic gastric cancer},
	volume = {24},
	issn = {1546-170X},
	doi = {10.1038/s41591-018-0101-z},
	language = {eng},
	number = {9},
	journal = {Nature Medicine},
	author = {Kim, Seung Tae and Cristescu, Razvan and Bass, Adam J. and Kim, Kyoung-Mee and Odegaard, Justin I. and Kim, Kyung and Liu, Xiao Qiao and Sher, Xinwei and Jung, Hun and Lee, Mijin and Lee, Sujin and Park, Se Hoon and Park, Joon Oh and Park, Young Suk and Lim, Ho Yeong and Lee, Hyuk and Choi, Mingew and Talasaz, AmirAli and Kang, Peter Soonmo and Cheng, Jonathan and Loboda, Andrey and Lee, Jeeyun and Kang, Won Ki},
	month = sep,
	year = {2018},
	pmid = {30013197},
	pages = {1449--1458},
}

@article{liu_integrative_2019,
	title = {Integrative molecular and clinical modeling of clinical outcomes to {PD1} blockade in patients with metastatic melanoma},
	volume = {25},
	copyright = {2019 The Author(s)},
	issn = {1546-170X},
	url = {https://www.nature.com/articles/s41591-019-0654-5},
	doi = {10.1038/s41591-019-0654-5},
	language = {en},
	number = {12},
	urldate = {2026-02-03},
	journal = {Nature Medicine},
	author = {Liu, David and Schilling, Bastian and Liu, Derek and Sucker, Antje and Livingstone, Elisabeth and Jerby-Arnon, Livnat and Zimmer, Lisa and Gutzmer, Ralf and Satzger, Imke and Loquai, Carmen and Grabbe, Stephan and Vokes, Natalie and Margolis, Claire A. and Conway, Jake and He, Meng Xiao and Elmarakeby, Haitham and Dietlein, Felix and Miao, Diana and Tracy, Adam and Gogas, Helen and Goldinger, Simone M. and Utikal, Jochen and Blank, Christian U. and Rauschenberg, Ricarda and von Bubnoff, Dagmar and Krackhardt, Angela and Weide, Benjamin and Haferkamp, Sebastian and Kiecker, Felix and Izar, Ben and Garraway, Levi and Regev, Aviv and Flaherty, Keith and Paschen, Annette and Van Allen, Eliezer M. and Schadendorf, Dirk},
	month = dec,
	year = {2019},
	pages = {1916--1927},
}

@article{mcdermott_clinical_2018,
	title = {Clinical activity and molecular correlates of response to atezolizumab alone or in combination with bevacizumab versus sunitinib in renal cell carcinoma},
	volume = {24},
	copyright = {2018 The Author(s)},
	issn = {1546-170X},
	url = {https://www.nature.com/articles/s41591-018-0053-3},
	doi = {10.1038/s41591-018-0053-3},
	language = {en},
	number = {6},
	urldate = {2026-02-03},
	journal = {Nature Medicine},
	author = {McDermott, David F. and Huseni, Mahrukh A. and Atkins, Michael B. and Motzer, Robert J. and Rini, Brian I. and Escudier, Bernard and Fong, Lawrence and Joseph, Richard W. and Pal, Sumanta K. and Reeves, James A. and Sznol, Mario and Hainsworth, John and Rathmell, W. Kimryn and Stadler, Walter M. and Hutson, Thomas and Gore, Martin E. and Ravaud, Alain and Bracarda, Sergio and Suárez, Cristina and Danielli, Riccardo and Gruenwald, Viktor and Choueiri, Toni K. and Nickles, Dorothee and Jhunjhunwala, Suchit and Piault-Louis, Elisabeth and Thobhani, Alpa and Qiu, Jiaheng and Chen, Daniel S. and Hegde, Priti S. and Schiff, Christina and Fine, Gregg D. and Powles, Thomas},
	month = jun,
	year = {2018},
	pages = {749--757},
}

@article{van_allen_genomic_2015,
	title = {Genomic correlates of response to {CTLA}-4 blockade in metastatic melanoma},
	volume = {350},
	copyright = {http://www.sciencemag.org/about/science-licenses-journal-article-reuse},
	issn = {0036-8075, 1095-9203},
	url = {https://www.science.org/doi/10.1126/science.aad0095},
	doi = {10.1126/science.aad0095},
	language = {en},
	number = {6257},
	urldate = {2026-02-03},
	journal = {Science},
	author = {Van Allen, Eliezer M. and Miao, Diana and Schilling, Bastian and Shukla, Sachet A. and Blank, Christian and Zimmer, Lisa and Sucker, Antje and Hillen, Uwe and Geukes Foppen, Marnix H. and Goldinger, Simone M. and Utikal, Jochen and Hassel, Jessica C. and Weide, Benjamin and Kaehler, Katharina C. and Loquai, Carmen and Mohr, Peter and Gutzmer, Ralf and Dummer, Reinhard and Gabriel, Stacey and Wu, Catherine J. and Schadendorf, Dirk and Garraway, Levi A.},
	month = oct,
	year = {2015},
	pages = {207--211},
}

@article{riaz_tumor_2017,
	title = {Tumor and {Microenvironment} {Evolution} during {Immunotherapy} with {Nivolumab}},
	volume = {171},
	issn = {0092-8674},
	url = {https://www.sciencedirect.com/science/article/pii/S0092867417311224},
	doi = {10.1016/j.cell.2017.09.028},
	number = {4},
	urldate = {2026-02-03},
	journal = {Cell},
	author = {Riaz, Nadeem and Havel, Jonathan J. and Makarov, Vladimir and Desrichard, Alexis and Urba, Walter J. and Sims, Jennifer S. and Hodi, F. Stephen and Martín-Algarra, Salvador and Mandal, Rajarsi and Sharfman, William H. and Bhatia, Shailender and Hwu, Wen-Jen and Gajewski, Thomas F. and Slingluff, Craig L. and Chowell, Diego and Kendall, Sviatoslav M. and Chang, Han and Shah, Rachna and Kuo, Fengshen and Morris, Luc G. T. and Sidhom, John-William and Schneck, Jonathan P. and Horak, Christine E. and Weinhold, Nils and Chan, Timothy A.},
	month = nov,
	year = {2017},
	pages = {934--949.e16},
}

@misc{eddy_cri_2020,
	title = {{CRI} {iAtlas}: an interactive portal for immuno-oncology research},
	copyright = {https://creativecommons.org/licenses/by/4.0/},
	shorttitle = {{CRI} {iAtlas}},
	url = {https://f1000research.com/articles/9-1028},
	doi = {10.12688/f1000research.25141.1},
	language = {en},
	urldate = {2026-01-27},
	publisher = {F1000Research},
	author = {Eddy, James A. and Thorsson, Vésteinn and Lamb, Andrew E. and Gibbs, David L. and Heimann, Carolina and Yu, Jia Xin and Chung, Verena and Chae, Yooree and Dang, Kristen and Vincent, Benjamin G. and Shmulevich, Ilya and Guinney, Justin},
	month = aug,
	year = {2020},
}

@article{rizvi_mutational_2015,
	title = {Mutational landscape determines sensitivity to {PD}-1 blockade in non–small cell lung cancer},
	volume = {348},
	copyright = {http://www.sciencemag.org/about/science-licenses-journal-article-reuse},
	issn = {0036-8075, 1095-9203},
	url = {https://www.science.org/doi/10.1126/science.aaa1348},
	doi = {10.1126/science.aaa1348},
	language = {en},
	number = {6230},
	urldate = {2025-08-21},
	journal = {Science},
	author = {Rizvi, Naiyer A. and Hellmann, Matthew D. and Snyder, Alexandra and Kvistborg, Pia and Makarov, Vladimir and Havel, Jonathan J. and Lee, William and Yuan, Jianda and Wong, Phillip and Ho, Teresa S. and Miller, Martin L. and Rekhtman, Natasha and Moreira, Andre L. and Ibrahim, Fawzia and Bruggeman, Cameron and Gasmi, Billel and Zappasodi, Roberta and Maeda, Yuka and Sander, Chris and Garon, Edward B. and Merghoub, Taha and Wolchok, Jedd D. and Schumacher, Ton N. and Chan, Timothy A.},
	month = apr,
	year = {2015},
	pages = {124--128},
}

@article{weinstein_cancer_2013,
	title = {The {Cancer} {Genome} {Atlas} {Pan}-{Cancer} {Analysis} {Project}},
	volume = {45},
	issn = {1061-4036},
	url = {https://www.ncbi.nlm.nih.gov/pmc/articles/PMC3919969/},
	doi = {10.1038/ng.2764},
	number = {10},
	urldate = {2025-08-08},
	journal = {Nature genetics},
	author = {Weinstein, John N. and Collisson, Eric A. and Mills, Gordon B. and Shaw, Kenna M. and Ozenberger, Brad A. and Ellrott, Kyle and Shmulevich, Ilya and Sander, Chris and Stuart, Joshua M.},
	month = oct,
	year = {2013},
	pmid = {24071849},
	pmcid = {PMC3919969},
	pages = {1113--1120},
}

@misc{shen_generalizable_2025,
	title = {Generalizable {AI} predicts immunotherapy outcomes across cancers and treatments},
	copyright = {http://creativecommons.org/licenses/by-nc/4.0/},
	url = {http://medrxiv.org/lookup/doi/10.1101/2025.05.01.25326820},
	doi = {10.1101/2025.05.01.25326820},
	language = {en},
	urldate = {2025-08-08},
	author = {Shen, Wanxiang and Nguyen, Thinh H. and Li, Michelle M. and Huang, Yepeng and Moon, Intae and Nair, Nitya and Marbach, Daniel and Zitnik, Marinka},
	month = may,
	year = {2025},
}

@article{kong_network-based_2022,
	title = {Network-based machine learning approach to predict immunotherapy response in cancer patients},
	volume = {13},
	copyright = {2022 The Author(s)},
	issn = {2041-1723},
	url = {https://www.nature.com/articles/s41467-022-31535-6},
	doi = {10.1038/s41467-022-31535-6},
	language = {en},
	number = {1},
	urldate = {2025-08-07},
	journal = {Nature Communications},
	author = {Kong, JungHo and Ha, Doyeon and Lee, Juhun and Kim, Inhae and Park, Minhyuk and Im, Sin-Hyeog and Shin, Kunyoo and Kim, Sanguk},
	month = jun,
	year = {2022},
	pages = {3703},
}

@article{rabinovich_immunosuppressive_2007,
	title = {Immunosuppressive {Strategies} that are {Mediated} by {Tumor} {Cells}},
	volume = {25},
	issn = {0732-0582, 1545-3278},
	url = {https://www.annualreviews.org/doi/10.1146/annurev.immunol.25.022106.141609},
	doi = {10.1146/annurev.immunol.25.022106.141609},
	language = {en},
	number = {1},
	urldate = {2024-12-02},
	journal = {Annual Review of Immunology},
	author = {Rabinovich, Gabriel A. and Gabrilovich, Dmitry and Sotomayor, Eduardo M.},
	month = apr,
	year = {2007},
	pages = {267--296},
}

@article{leach_enhancement_1996,
	title = {Enhancement of {Antitumor} {Immunity} by {CTLA}-4 {Blockade}},
	volume = {271},
	issn = {0036-8075, 1095-9203},
	url = {https://www.science.org/doi/10.1126/science.271.5256.1734},
	doi = {10.1126/science.271.5256.1734},
	language = {en},
	number = {5256},
	urldate = {2024-12-02},
	journal = {Science},
	author = {Leach, Dana R. and Krummel, Matthew F. and Allison, James P.},
	month = mar,
	year = {1996},
	pages = {1734--1736},
}

@article{simoni_bystander_2018,
	title = {Bystander {CD8}+ {T} cells are abundant and phenotypically distinct in human tumour infiltrates},
	volume = {557},
	copyright = {2018 Macmillan Publishers Ltd., part of Springer Nature},
	issn = {1476-4687},
	url = {https://www.nature.com/articles/s41586-018-0130-2},
	doi = {10.1038/s41586-018-0130-2},
	language = {en},
	number = {7706},
	urldate = {2024-12-02},
	journal = {Nature},
	author = {Simoni, Yannick and Becht, Etienne and Fehlings, Michael and Loh, Chiew Yee and Koo, Si-Lin and Teng, Karen Wei Weng and Yeong, Joe Poh Sheng and Nahar, Rahul and Zhang, Tong and Kared, Hassen and Duan, Kaibo and Ang, Nicholas and Poidinger, Michael and Lee, Yin Yeng and Larbi, Anis and Khng, Alexis J. and Tan, Emile and Fu, Cherylin and Mathew, Ronnie and Teo, Melissa and Lim, Wan Teck and Toh, Chee Keong and Ong, Boon-Hean and Koh, Tina and Hillmer, Axel M. and Takano, Angela and Lim, Tony Kiat Hon and Tan, Eng Huat and Zhai, Weiwei and Tan, Daniel S. W. and Tan, Iain Beehuat and Newell, Evan W.},
	month = may,
	year = {2018},
	pages = {575--579},
}

@article{garon_pembrolizumab_2015,
	title = {Pembrolizumab for the {Treatment} of {Non}–{Small}-{Cell} {Lung} {Cancer}},
	volume = {372},
	issn = {0028-4793, 1533-4406},
	url = {http://www.nejm.org/doi/10.1056/NEJMoa1501824},
	doi = {10.1056/NEJMoa1501824},
	language = {en},
	number = {21},
	urldate = {2024-12-02},
	journal = {New England Journal of Medicine},
	author = {Garon, Edward B. and Rizvi, Naiyer A. and Hui, Rina and Leighl, Natasha and Balmanoukian, Ani S. and Eder, Joseph Paul and Patnaik, Amita and Aggarwal, Charu and Gubens, Matthew and Horn, Leora and Carcereny, Enric and Ahn, Myung-Ju and Felip, Enriqueta and Lee, Jong-Seok and Hellmann, Matthew D. and Hamid, Omid and Goldman, Jonathan W. and Soria, Jean-Charles and Dolled-Filhart, Marisa and Rutledge, Ruth Z. and Zhang, Jin and Lunceford, Jared K. and Rangwala, Reshma and Lubiniecki, Gregory M. and Roach, Charlotte and Emancipator, Kenneth and Gandhi, Leena},
	month = may,
	year = {2015},
	pages = {2018--2028},
}

@article{drake_breathing_2014,
	title = {Breathing new life into immunotherapy: review of melanoma, lung and kidney cancer},
	volume = {11},
	copyright = {2013 Springer Nature Limited},
	issn = {1759-4782},
	shorttitle = {Breathing new life into immunotherapy},
	url = {https://www.nature.com/articles/nrclinonc.2013.208},
	doi = {10.1038/nrclinonc.2013.208},
	language = {en},
	number = {1},
	urldate = {2024-12-02},
	journal = {Nature Reviews Clinical Oncology},
	author = {Drake, Charles G. and Lipson, Evan J. and Brahmer, Julie R.},
	month = jan,
	year = {2014},
	pages = {24--37},
}

@article{grivennikov_immunity_2010,
	title = {Immunity, {Inflammation}, and {Cancer}},
	volume = {140},
	issn = {0092-8674},
	url = {https://www.sciencedirect.com/science/article/pii/S0092867410000607},
	doi = {10.1016/j.cell.2010.01.025},
	number = {6},
	urldate = {2024-12-02},
	journal = {Cell},
	author = {Grivennikov, Sergei I. and Greten, Florian R. and Karin, Michael},
	month = mar,
	year = {2010},
	pages = {883--899},
}

@article{li_informing_2024,
	title = {Informing immunotherapy with multi-omics driven machine learning},
	volume = {7},
	issn = {2398-6352},
	url = {https://www.nature.com/articles/s41746-024-01043-6},
	doi = {10.1038/s41746-024-01043-6},
	language = {en},
	number = {1},
	urldate = {2024-11-19},
	journal = {npj Digital Medicine},
	author = {Li, Yawei and Wu, Xin and Fang, Deyu and Luo, Yuan},
	month = mar,
	year = {2024},
	pages = {67},
}
\bibliographystyle{mlgenx_conference}

\appendix
\section{Appendix}
\subsection{Data}
\subsubsection{Cohorts}
\label{app:cohorts}
The cohorts used for LOCO evaluation are given in Table \ref{tab:cohorts}.
% \usepackage{booktabs}
\begin{table}[hbtp]
\centering
\caption{Cohorts used for LOCO evaluation along with the cancer type, drug used, number of samples and size category as well as a citation to the publication.\newline}
\label{tab:cohorts}
\begin{tabular}{@{}lllll@{}}
\toprule
\textbf{Cohort}           & \textbf{Cancer Type} & \textbf{Drug} & \textbf{\begin{tabular}[c]{@{}l@{}}Number of\\ Samples\end{tabular}} & \textbf{Size Category}   \\ \midrule
\citet{mariathasan_tgfb_2018}                & BLCA                 & Atezolizumab  & 298                                                                  & Large (\textgreater{}50) \\
\citet{mcdermott_clinical_2018}               & KIRC                 & Atezolizumab  & 247                                                                  & Large (\textgreater{}50) \\
\citet{liu_integrative_2019}         & SKCM                 & Nivolumab     & 121                                                                  & Large (\textgreater{}50) \\
\citet{gide_distinct_2019}          & SKCM                 & Nivolumab     & 73                                                                   & Large (\textgreater{}50) \\
\citet{riaz_tumor_2017}     & SKCM                 & Nivolumab     & 49                                                                   & Small (\textless{}50)    \\
\citet{kim_comprehensive_2018}         & STAD                 & Pembrolizumab & 45                                                                   & Small (\textless{}50)    \\
\citet{van_allen_genomic_2015} & SKCM                 & Ipilimumab    & 41                                                                   & Small (\textless{}50)    \\
\citet{hugo_genomic_2016}       & SKCM                 & Pembrolizumab & 26                                                                   & Small (\textless{}50)    \\ \bottomrule
\end{tabular}
\end{table}

\subsubsection{Preprocessing}
The following preprocessing was applied to gene expression data on CRI iAtlas. ENST counts were generated by trimming FASTQ reads with TrimGalore (v0.6.2), aligning them to GRCh38 (gtf: v103; ref: p13) using STAR (v2.7.0f), and performing quantification with Salmon (v1.1.0)
\subsection{Model}
\label{app:model}

\subsubsection{Pathway Consistency}
\begin{equation}
    \mathcal{L}_{\text{pathway}} = \frac{1}{B} \sum_{i=1}^{B} \|\mathbf{p}_i - f_{\text{path}}(\text{mean}(\mathbf{E}_i))\|_2^2 
\end{equation}
where $\mathbf{E}i \in \mathbb{R}^{L \times d_e}$ is the gene encoding for sample $i$, $L$ is the number of gene tokens, $d_e$ is the dimension of gene encoding, $f_{\text{path}}$ is the pathway predictor head, $\mathbf{p}_i \in \mathbb{R}^{42}$ are target pathway scores and $B$ is the batch size.

\subsubsection{Concept Alignment}
\begin{equation}
    \mathcal{L}_{\text{align}} = \|\mathbf{C} W - \mathbf{B}\|_2^2
\end{equation}
where $\mathbf{C} \in \mathbb{R}^{B \times 44}$ are concepts, $W \in \mathbb{R}^{44 \times d_b}$ is a learnable projection, $\mathbf{B} \in \mathbb{R}^{B \times d_b}$ are biomarkers, $B$ is the batch size and $d_b$ is the number of biomarkers. 44 is the number of biological concepts generated by COMPASS concept bottleneck.

\subsubsection{Auxiliary Tasks}
\begin{equation}
    \mathcal{L}_{\text{aux}} = \sum_{k \in \{\text{TIDE}, \text{IPRES}, \text{pheno}\}} \frac{1}{B} \sum_{i=1}^{B} \|\mathbf{t}_i^k - f_k(\mathbf{c}_i)\|_2^2
\end{equation}
where $f_k$ is the auxiliary decoder head for task $k$, $\mathbf{c}_i \in \mathbb{R}^{44}$ are concepts, and $\mathbf{t}_i^k$ are target scores for task $k$.

\subsubsection{Treatment Gating}
\begin{equation}
    \mathbf{g} = \sigma(W_2 \cdot \text{ReLU}(W_1 \cdot \mathbf{e}_t + b_1) + b_2)
    \quad
    \mathbf{c}' = \mathbf{c} \odot \mathbf{g}
\end{equation}

where $\mathbf{e}_t \in \mathbb{R}^{d_h}$ is the treatment embedding, $\mathbf{g} \in \mathbb{R}^{44}$ are gating weights, $\mathbf{c} \in \mathbb{R}^{44}$ are biological concepts, and $\odot$ denotes element-wise multiplication.

\subsection{Results}
\subsubsection{Ablation Study}
\label{app:ablation}
We conducted an ablation study to understand how generalisability changes with the absence of one component at a time and hence find out the most influential components.
\begin{table}[hbtp]
\centering
\caption{Ablation study of various components in BioCOMPASS. Each row shows performance when one component is disabled. Values are mean ± 95\% CI margin across five random seeds across all cohorts in LOCO setting. Treatment gating contributes most to performance improvement, followed by pathway consistency. Lower is better here as it shows that taking way that component reduced performance the most.\newline}
\begin{tabular}{@{}llllll@{}}
\toprule
\textbf{Config} & \textbf{Acc (\%)}     & \textbf{AUC (\%)}     & \textbf{F1 (\%)}      & \textbf{Prec (\%)}    & \textbf{Recall (\%)}  \\ \midrule
No Gating       & \textbf{72.13 ± 3.32} & \textbf{64.16 ± 4.52} & \textbf{48.23 ± 5.07} & \textbf{51.07 ± 6.48} & 58.65 ± 8.37          \\
No Pathway      & 72.24 ± 2.88          & 67.53 ± 3.51          & 52.05 ± 3.74          & 53.96 ± 5.88          & \textbf{57.01 ± 5.52} \\
No Auxiliary    & 72.55 ± 3.13          & 68.32 ± 3.06          & 51.94 ± 4.04          & 53.58 ± 5.36          & 57.10 ± 6.15          \\
No Alignment    & 72.52 ± 3.02          & 68.09 ± 3.48          & 52.50 ± 4.20          & 54.78 ± 5.76          & 58.83 ± 6.35          \\ \bottomrule
\end{tabular}
\end{table}

\subsubsection{Additional generalisability experiments}
\label{app:addgen}
\begin{figure}[htbp]
    \centering
    \includegraphics[width=1.0\linewidth]{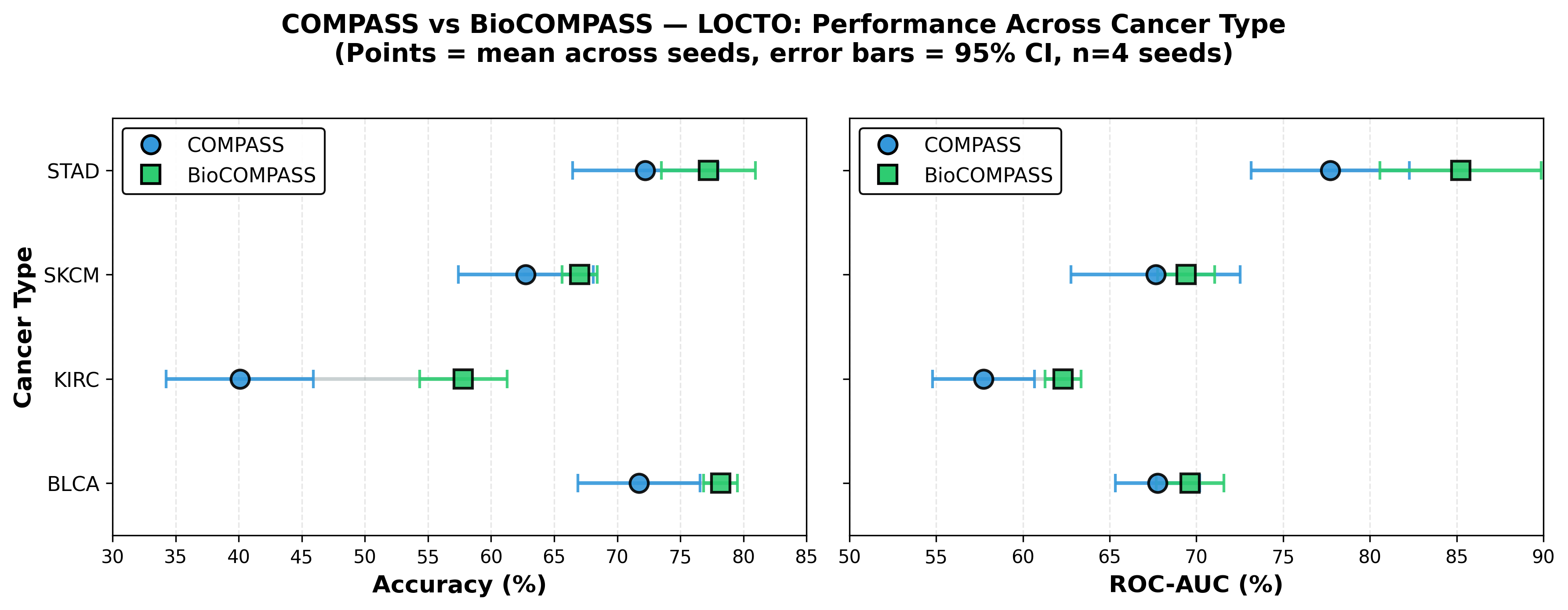}
    \caption{Points show mean performance across four random seeds with 95\% CI error bars. Circles: COMPASS; Squares: BioCOMPASS. Please note that the subplots have different scales on x-axis.}
    \label{fig:locto}
\end{figure}

\begin{figure}[htbp]
    \centering
    \includegraphics[width=1.0\linewidth]{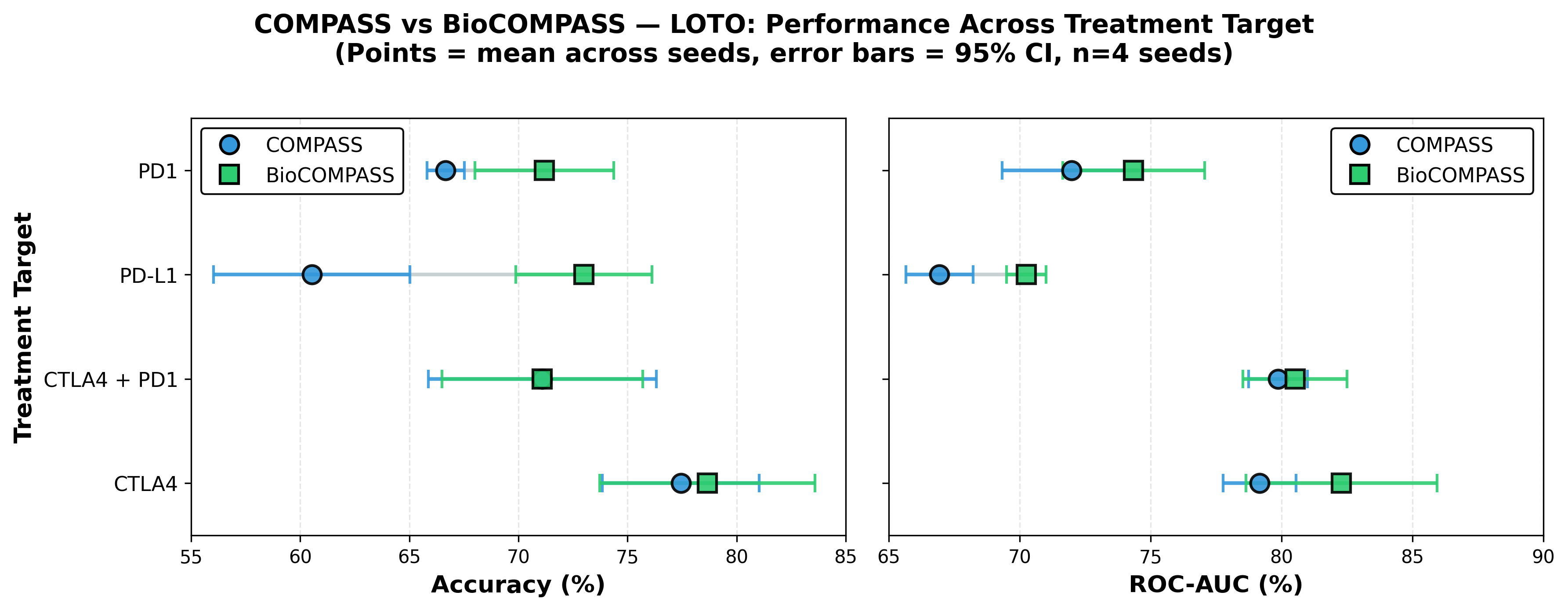}
    \caption{Points show mean performance across four random seeds with 95\% CI error bars. Circles: COMPASS; Squares: BioCOMPASS. Please note that the subplots have different scales on x-axis.}
    \label{fig:loto}
\end{figure}

The results of additional experiments in LOCTO and LOTO settings are in Table \ref{tab:addgen}. The LOCTO setting involves leaving one of four types of cancer at a time: bladder urothelial carcinoma (BLCA), renal clear cell carcinoma (KIRC), cutaneous Cutaneous melanoma (SKCM), and Stomach adenocarcinoma (STAD). The LOTO setting includes leaving out one of four immunotherapy treatment targets: PD-1, PD-L1, CTLA-4 and CTLA-4 + PD-1. BioCOMPASS performs better in both settings in all metrics except recall which could be because BioCOMPASS might be more conservative in its predictions, as explained earlier. Figures \ref{fig:locto} and \ref{fig:loto} show average performance on each left-out cancer type and treatment target respectively across 4 seeds.

\begin{table}[htbp]
\centering
\caption{Performance of COMPASS (C) and BioCOMPASS (BC) in LOCTO and LOTO settings. This table shows the average performance (in \%) across 4 left-out cancer types (BLCA, KIRC, SKCM, STAD) and 4 immunotherapy treatment targets (PD-1, PD-L1, CTLA-4, CTLA-4 + PD-1). The 95\% confidence intervals (CI) show variation across 4 seeds. \newline}
\label{tab:addgen}
\begin{tabular}{@{}lllllll@{}}
\toprule
\textbf{Setting}       & \textbf{Model} & \textbf{Accuracy}     & \textbf{ROC AUC}      & \textbf{F1}           & \textbf{Precision}    & \textbf{Recall}       \\ \midrule
\multirow{2}{*}{LOCTO} & C              & 61.69 ± 6.99          & 67.72 ± 3.87          & 42.41 ± 5.40          & 42.61 ± 13.69         & \textbf{51.50 ± 8.44} \\
                       & BC             & \textbf{70.05 ± 1.63} & \textbf{71.65 ± 3.02} & \textbf{49.16 ± 5.72} & \textbf{51.66 ± 2.38} & 50.37 ± 10.22         \\
\multirow{2}{*}{LOTO}  & C              & 68.93 ± 2.95          & 74.49 ± 1.56          & 57.02 ± 2.16          & 55.44 ± 2.04          & \textbf{63.53 ± 4.86} \\
                       & BC             & \textbf{73.49 ± 1.36} & \textbf{76.85 ± 3.56} & \textbf{60.53 ± 2.66} & \textbf{61.01 ± 1.84} & 63.36 ± 4.28          \\ \bottomrule
\end{tabular}
\end{table}

\subsection{Baseline Methods}
\label{app:baseline}
We trained a logistic regression model on  biomarker-based baseline methods to analyse their generalisability. The implementation from COMPASS was used for this purpose and are described in Table \ref{tab:baseline_desc}. We also trained logistic regression and other standard machine learning methods on gene expression data and biomarker data. Results are in Table \ref{tab:baseline}. As can be seen from the results, these methods fail to generalise across unseen test groups.

\begin{table}[htbp]
\centering
\caption{Description of biomarker-based baseline methods. These were used based on the implementation by COMPASS.\newline}
\label{tab:baseline_desc}
\begin{tabular}{@{}lll@{}}
\toprule
\textbf{Method} & \textbf{Description}                                                                                            & \textbf{Reference}                                                                                      \\ \midrule
GeneBio         & \begin{tabular}[c]{@{}l@{}}Combined score of immunotherapy target markers\\ PD1/PDL1/CTLA4\end{tabular}         & \citet{kong_network-based_2022}                                                                         \\
CTLA4           & Expression of CTLA4 as a single ICI target marker                                                               & \citet{kong_network-based_2022}                                                                         \\
PD1             & Expression of PDCD1 as a single ICI target marker                                                               & \citet{kong_network-based_2022}                                                                         \\
PDL1            & Expression of CD274 as a single ICI target marker                                                               & \citet{kong_network-based_2022}                                                                         \\
CD8             & \begin{tabular}[c]{@{}l@{}}CD8$^+$ T cell score derived from average expression of\\ CD8A and CD8B\end{tabular} & \begin{tabular}[c]{@{}l@{}}\citet{chen_analysis_2016}\\ \citet{kong_network-based_2022}\end{tabular}    \\
CIS             & \begin{tabular}[c]{@{}l@{}}Cytotoxic immune signature score averaging cytotoxic\\ immune genes\end{tabular}     & \citet{davoli_tumor_2017}                                                                               \\
Teff            & T-effector/IFN-$\gamma$ signature score averaging T-effector genes                                              & \citet{fehrenbacher_atezolizumab_2016}                                                                  \\
PGM             & \begin{tabular}[c]{@{}l@{}}Prognostic gene-pair model; top pair:lymphocyte MAP4K1\\ and tumor TBX3\end{tabular} & \citet{freeman_combined_2022}                                                                           \\
NRS             & \begin{tabular}[c]{@{}l@{}}Neoadjuvant response signature score averaging NRS gene\\ expression\end{tabular}    & \citet{huang_single_2019}                                                                               \\
IFNG            & IFN-$\gamma$ response score based on an 18-gene signature                                                       & \citet{ayers_ifn-grelated_2017}                                                                          \\
IMPRES          & \begin{tabular}[c]{@{}l@{}}Sum of expression ratios across 15 immune/checkpoint gene\\ pairs\end{tabular}       & \citet{auslander_robust_2018}                                                                           \\
TIDE            & Tumor Immune Dysfunction and Exclusion composite score                                                          & \citet{jiang_signatures_2018}                                                                           \\
CTL             & \begin{tabular}[c]{@{}l@{}}Cytotoxic T lymphocyte score averaging CTL signature gene\\ expression\end{tabular}  & \citet{jiang_signatures_2018}                                                                           \\
TAM             & \begin{tabular}[c]{@{}l@{}}Tumor-associated macrophage score averaging TAM signature\\ genes\end{tabular}       & \begin{tabular}[c]{@{}l@{}}\citet{joyce_t_2015}\\ \citet{jiang_signatures_2018}\end{tabular}            \\
Texh            & T-cell exhaustion score averaging exhaustion signature genes                                                    & \begin{tabular}[c]{@{}l@{}}\citet{giordano_molecular_2015}\\ \citet{jiang_signatures_2018}\end{tabular} \\
CKS             & \begin{tabular}[c]{@{}l@{}}12-chemokine signature score using PC1 of chemokine gene\\ expression\end{tabular}   & \citet{messina_12-chemokine_2012}                                                                       \\
CAF             & \begin{tabular}[c]{@{}l@{}}Cancer-associated fibroblast score averaging CAF signature\\ genes\end{tabular}      & \citet{nurmik_search_2020}                                                                              \\
IS              & \begin{tabular}[c]{@{}l@{}}Immune score averaging expression of a panel of immune\\ genes\end{tabular}          & \citet{roh_integrated_2017}                                                                             \\
ICA             & \begin{tabular}[c]{@{}l@{}}Immune cytolytic activity score based on GZMA and PRF1\\ expression\end{tabular}     & \citet{rooney_molecular_2015}                                                                           \\
MIAS            & \begin{tabular}[c]{@{}l@{}}MHC-I association immune score computed via ssGSEA on\\ 100 genes\end{tabular}       & \citet{wu_prediction_2022}                                                                              \\
GEP             & \begin{tabular}[c]{@{}l@{}}T cell-inflamed gene expression profile score via ssGSEA on\\ 18 genes\end{tabular}  & \begin{tabular}[c]{@{}l@{}}\citet{cristescu_pan-tumor_2018}\\ \citet{wu_prediction_2022}\end{tabular}   \\
NetBio          & \begin{tabular}[c]{@{}l@{}}Score derived from the top 200 ICI target-proximal network\\ genes\end{tabular}      & \citet{kong_network-based_2022}                                                                         \\ \bottomrule
\end{tabular}
\end{table}

\begin{table}[htbp]
\centering
\caption{Performance of baseline methods in LOCO (C), LOCTO (CT) and LOTO (T) settings. It can be seen that all baseline methods show poor generalisability. Immune score based methods were used based on the implementation by COMPASS. LR: Logistic regression, GBM: Gradient boosting machine, RF: Random forest, PCA: Principal component analysis\newline}
\label{tab:baseline}
\begin{tabular}{@{}lccccccccc@{}}
\toprule
\multirow{2}{*}{\textbf{Method}} & \multicolumn{3}{c}{\textbf{AUC (\%)}}            & \multicolumn{3}{c}{\textbf{Accuracy (\%)}}       & \multicolumn{3}{c}{\textbf{F1 (\%)}}             \\ \cmidrule(l){2-10} 
                                 & \textbf{C}     & \textbf{CT}    & \textbf{T}     & \textbf{C}     & \textbf{CT}    & \textbf{T}     & \textbf{C}     & \textbf{CT}    & \textbf{T}     \\ \midrule
\multicolumn{10}{l}{\textbf{Logistic regression on baseline methods}}                                                                                                                     \\ \midrule
CKS                              & 61.88          & 63.96          & 63.05          & 57.70          & 59.99          & 59.30          & 48.23          & \textbf{47.95} & 48.78          \\
GEP                              & \textbf{62.83} & 65.50          & \textbf{65.42} & 57.27          & 57.05          & 60.02          & \textbf{48.30} & 46.60          & 47.77          \\
IFNG                             & 61.97          & 64.82          & 62.89          & 65.57          & 68.51          & 65.23          & 38.85          & 38.36          & 39.31          \\
CD8                              & 61.13          & 64.39          & 61.99          & 57.27          & 59.74          & 62.65          & 44.61          & 42.89          & 48.80          \\
Teff                             & 62.39          & \textbf{66.06} & 63.98          & 56.71          & 60.33          & 63.70          & 44.04          & 46.35          & 39.79          \\
IS                               & 62.54          & 64.89          & 62.39          & 58.73          & 62.16          & 60.57          & 40.46          & 42.62          & 43.64          \\
ICA                              & 60.40          & 63.23          & 62.41          & 60.45          & 63.09          & 63.95          & 40.00          & 41.96          & 41.50          \\
PDL1                             & 59.72          & 62.92          & 59.71          & 59.24          & 60.59          & 58.40          & 42.13          & 46.94          & 45.76          \\
CTL                              & 61.21          & 64.35          & 63.61          & 60.83          & 64.95          & 62.16          & 37.55          & 39.25          & 35.84          \\
CIS                              & 60.81          & 62.98          & 61.59          & 64.56          & 67.81          & 53.65          & 36.12          & 40.75          & 41.41          \\
GeneBio                          & 60.75          & 64.20          & 61.57          & 57.31          & 62.69          & 61.59          & 39.04          & 41.11          & 40.67          \\
CTLA4                            & 59.53          & 60.37          & 62.16          & 58.19          & 57.07          & 55.69          & 46.12          & 40.73          & \textbf{48.96} \\
PD1                              & 60.43          & 60.93          & 59.13          & 56.79          & 56.67          & 58.70          & 45.00          & 42.37          & 47.32          \\
MIAS                             & 61.09          & 62.22          & 60.55          & 62.98          & 66.46          & 54.25          & 38.08          & 41.80          & 39.30          \\
TAM                              & 55.62          & 58.87          & 48.53          & 56.28          & 64.40          & 54.32          & 44.67          & 40.52          & 45.67          \\
NRS                              & 56.96          & 58.37          & 61.28          & 56.93          & 58.92          & 52.64          & 37.35          & 41.81          & 40.51          \\
PGM                              & 59.27          & 57.42          & 58.50          & 57.64          & 63.07          & 58.48          & 35.59          & 29.66          & 43.15          \\
IMPRES                           & 56.01          & 52.20          & 57.19          & 54.41          & 52.39          & 53.73          & 32.98          & 36.74          & 41.63          \\
NetBio                           & 56.36          & 50.01          & 53.98          & 56.52          & 45.41          & 53.71          & 30.58          & 33.69          & 39.69          \\
Texh                             & 49.14          & 48.05          & 45.56          & 43.67          & 45.06          & 46.20          & 45.93          & 40.51          & 48.89          \\
CAF                              & 54.45          & 56.73          & 41.94          & 45.04          & 41.09          & 42.10          & 43.24          & 41.22          & 39.56          \\
TIDE                             & 47.61          & 53.46          & 41.63          & 59.48          & 62.18          & 46.90          & 29.81          & 30.54          & 27.57          \\ \midrule
\multicolumn{10}{l}{\textbf{Standard ML models on biomarkers}}                                                                                                                            \\ \midrule
LR                               & 59.61          & 61.85          & 57.14          & 58.95          & 62.85          & 54.26          & 40.57          & 39.04          & 41.27          \\
GBM                              & 53.04          & 52.44          & 60.10          & 55.96          & 61.39          & 60.97          & 33.28          & 31.01          & 40.75          \\
RF                               & 58.30          & 58.18          & 59.75          & 65.68          & 70.27          & 65.17          & 6.10           & 8.23           & 21.38          \\ \midrule
\multicolumn{10}{l}{\textbf{Standard ML models on gene expression}}                                                                                                                       \\ \midrule
PCA + LR                         & 59.45          & 59.82          & 60.84          & 59.70          & 65.85          & 53.39          & 33.18          & 24.74          & 46.35          \\
PCA + GBM                        & 55.07          & 52.82          & 62.64          & 60.43          & 65.80          & 63.27          & 30.63          & 20.88          & 36.75          \\
PCA + RF                         & 55.60          & 54.22          & 58.20          & \textbf{66.50} & \textbf{70.55} & \textbf{65.67} & 0.95           & 0.40           & 9.95           \\
LR                               & 55.32          & 57.58          & 54.49          & 60.63          & 62.25          & 62.75          & 24.18          & 28.91          & 33.52          \\
RF                               & 56.67          & 57.82          & 55.47          & 65.62          & 70.10          & 63.83          & 2.29           & 2.18           & 12.33          \\ \bottomrule
\end{tabular}
\end{table}

\subsection{Glossary}
\subsubsection{Immunotherapy \& Treatment}

\begin{description}
\item[PD-1 (Programmed Death 1)] A checkpoint protein that regulates immune responses; target for immunotherapy.

\item[PD-L1 (Programmed Death-Ligand 1)] A protein that binds to PD-1; its expression level is used as a biomarker.

\item[CTLA-4 (Cytotoxic T Lymphocyte-Associated Protein 4)] An immune checkpoint protein that downregulates immune responses; target for immunotherapy.

\item[Anti-PD-1 therapy] Treatment that blocks PD-1 (e.g., Nivolumab, Pembrolizumab).

\item[Anti-CTLA-4 therapy] Treatment that blocks CTLA-4 (e.g., Ipilimumab).

\item[Atezolizumab] An anti-PD-L1 immunotherapy drug.

\item[Nivolumab] An anti-PD-1 immunotherapy drug.

\item[Pembrolizumab] An anti-PD-1 immunotherapy drug.

\item[Ipilimumab] An anti-CTLA-4 immunotherapy drug.
\end{description}

\subsubsection{Cell Types \& Immune Components}

\begin{description}
\item[Cytotoxic T-cell (Cytotoxic T lymphocyte)] Immune cells that kill cancer cells.

\item[Plasma cell] B cells that produce antibodies.

\item[Immune cell infiltration] The presence of immune cells within the tumour.

\end{description}

\subsubsection{Biomarkers \& Scoring Methods}

\begin{description}
\item[TIDE] Tumour Immune Dysfunction and Exclusion score; biomarker for immunotherapy response.

\item[IPRES] Innate anti-PD-1 Resistance signature.

\item[Immune phenotypes] Classifications of tumours based on immune cell composition.

\item[Cell type abundances] Quantification of different immune cell populations in tumours.

\item[Pathway activity scores] Measurements of biological pathway activation (e.g., CTLA-4/PD-1 pathways).
\end{description}

\end{document}